\pdfoutput=1

\documentclass[11pt]{article}

\usepackage{times}
\usepackage{latexsym}
\usepackage[hyperref]{emnlp2021}
\usepackage{times}
\usepackage{latexsym}
\usepackage{graphicx} 
\usepackage{booktabs} 
\usepackage{placeins} 
\usepackage{microtype} 
\usepackage{enumitem} 
\usepackage{amsmath, bm} 

\usepackage[T1]{fontenc}

\usepackage[utf8]{inputenc}

\usepackage{microtype}

\title{Instructions for EMNLP 2021 Proceedings}

\title{Estimation of Summary-to-Text Inconsistency by Mismatched Embeddings}

\author{Oleg Vasilyev, John Bohannon \\
  Primer Technologies Inc. \\
  San Francisco, California \\
  \texttt{{oleg,john}@primer.ai}\\}

\begin{document}
\maketitle
\begin{abstract}
We propose a new reference-free summary quality evaluation measure, with emphasis on the faithfulness. The measure is designed to find and count all possible minute inconsistencies of the summary with respect to the source document. The proposed ESTIME, Estimator of Summary-to-Text Inconsistency by Mismatched Embeddings, correlates with expert scores in summary-level SummEval dataset stronger than other common evaluation measures not only in Consistency but also in Fluency. 
We also introduce a method of generating subtle factual errors in human summaries. We show that ESTIME is more sensitive to subtle errors than other common evaluation measures.
\end{abstract}

\section{Introduction}
Summarization must preserve the factual consistency of the summary with the text. Human annotation of factual consistency can be accompanied with detailed classification of factual errors, thus giving a hope that the annotation scores are reasonably objective \cite{Wojciech2020Evaluating, Dandan2020What, Oleg2020Sensitivity, Saadia2020GoFigure, Maynez2020Faithfulness}.

Factual consistency, or 'faithfulness' of a summary is one of several summary qualities; for the purpose of human annotation these qualities can be specified in different ways  \cite{Stratos2019Sum, Wojciech2020Evaluating, Lisa2018Robust, Oleg2020Sensitivity, Alexander2020SummEval}. Summarization models nowadays create satisfactorily fluent, coherent and informative summaries, but the factual consistency has a lot of room for improvement. Some factual errors (swapped named entities and crude hallucinations) are easily noticeable and make a summary look very bad right away; other factual errors could be hardly noticeable even for annotators \cite{Lux2020Truth, Oleg2020Sensitivity} - which is arguably even worse.
 
Existing summary evaluation measures are based on several approaches, which may be more sensitive to one or another quality.
A question-answering based evaluation estimates how helpful is the summary in answering questions about the source text \cite{Stratos2019SUMQE, Matan2019APES, Thomas2019SummaQA, Daniel2020Towards, Esin2020FEQA, Wang2020Asking}. A text reconstruction approach estimates how helpful is the summary in guessing parts of the source text \cite{Oleg2020BLANC, Oleg2020Sensitivity, Egan2021Play}. Evaluation measures that use some kind of text similarity can estimate how similar is the summary to special human-written reference summaries \cite{Tianyi2020BERTScore, Wei2020MoverScore, Lin2004ROUGE}, or, more realistically, how similar is the summary to the source text \cite{Yang2020SUPERT, Annie2009Automatically}.

In order to assess how well the summary factual consistency is evaluated, it is necessary either to have a dataset of human-annotated imperfect machine-generated summaries \cite{Manik2020ReEvaluating, Alexander2020SummEval}, or to have a dataset of artificially introduced factual errors in originally factually correct human-written summaries \cite{Wojciech2020Evaluating}.  
 \begin{figure*}[t]
     \centering
     \begin{tabular}{l}
     \toprule
     Given: $summary$; $text$; $model$, $tokenizer$; parameters $W=450$, $L=8$, $M=50$, $H=24$\\
     \\
     $tokens\_text = tokenizer.tokenize(text)$ \\
     $tokens\_summary = tokenizer.tokenize(summary)$ \\
     \\
     \# Get all embeddings from the text: \\
     Initialize $embeddings\_text$ of the same length as $tokens\_text$ \\
     While there are $text$ tokens with embeddings not taken: \\
     \hspace{.4cm}Select leftmost token $t$ with embedding not yet taken \\
     \hspace{.4cm}Make input widow of size $W$ start to the left by $M$ tokens from the index of $t$ \\
     \hspace{.4cm}Starting from token $t$, in the input window mask each $L$th token with embedding not yet taken\\
     \hspace{.4cm}Take the input by $model$, take $H$th hidden layer embeddings for all the masked tokens \\
     \hspace{.4cm}Place the obtained embeddings into their corresponding locations in $embeddings\_text$ \\
     \\
     \# Get all embeddings from the summary: \\
     Repeat the above for the $summary$. Result: $embeddings\_summary$ \\ 
     \\
     \# Count summary-to-text inconsistencies: \\
     $num\_inconsistencies = 0$ \\
     For each embedding $E_s$ in $embeddings\_summary$: \\
     \hspace{.4cm}Find in $embeddings\_text$ an embedding $E_t$ with highest similarity (dot-product) to $E_s$. \\
     \hspace{.4cm}If the tokens corresponding to $E_s$ and $E_t$ are not equal, then $num\_inconsistencies \mathrel{{+}{=}} 1$ \\
     \bottomrule
     \end{tabular}
     \caption{ESTIME: Estimation of Summary-to-Text Inconsistency by Mismatched Embeddings.}
     \label{fig:Algo_ESTIME}
 \end{figure*}

In this paper we focus on presenting a new evaluation measure with emphasis on factual consistency. Our contribution:
\begin{enumerate}[topsep=0pt,itemsep=-1ex,partopsep=1ex,parsep=1ex]
    \item We introduce ESTIME, Estimator of Summary-to-Text Inconsistency by Mismatched Embeddings. Using human-annotated machine-generated summaries from the datset SummEval \cite{Alexander2020SummEval}, we compare ESTIME with other common or promising evaluation measures.
    \item We introduce a natural no-heuristics method of generating subtle factual errors. We use it here to compare the performance of ESTIME with other measures on human-written summaries with generated subtle errors.  
\end{enumerate}

\section{Methods}
In order to estimate consistency of a summary with the text, we attempt to find all the summary tokens that could be related to a factual error. Our motivation is that transformer-made token embeddings are highly contextual \cite{Kawin2019Contextual}. We assume that even if phrased somewhat differently in the summary, the context should suggest approximately the same token embedding as an embedding that would be suggested by the corresponding context in the text. Thus, we check embeddings of all tokens of the summary that have one or more occurrences in the text. For each token embedding we find its match: the most similar embedding in the text. We assume that if the summary is factually correct then the matched tokens are the same. If the tokens are not the same, we count such mismatch as an indicator of inconsistency of the summary to the text. The algorithm for ESTIME is shown in more detail in Figure \ref{fig:Algo_ESTIME}.

This approach is different from matching similar embeddings for sake of measuring similarity (e.g. similarity between a summary and a reference summary in BERTScore \cite{Tianyi2020BERTScore}). It also differs from using a model trained to replace wrong tokens with correct ones \cite{Meng2020Factual, Wojciech2017Evaluating}. 

In preliminary evaluations, similar to the ones presented in the next sections, we have not found any improvement from adding up one or another flavour of heuristics, such as covering only named entities or only certain parts of speech. For finding most similar embedding we are using simple unnormalized dot-product, - replacing it by cosine similarity makes all the results presented in the next sections slightly worse. The embeddings are taken using the pretrained BERT model \cite{Jacob2018BERT} bert-large-uncased-whole-word-masking of transformers library \cite{Wolf2020Transformers}. While there is no crucial difference with other varieties of BERT, ALBERT and RoBERTa, this model showed a better overall performance, and we used it for evaluations in the next sections. 
We found that there is no strong dependency on the parameters $W$, $L$, $M$ of ESTIME, we set the input size $W=450$ close to maximal BERT input length, and the distances $L=8$ and $M=50$ reasonably large. We present results for $H=12$ as ESTIME-12, and for $H=24$ as ESTIME-24, corresponding to the embeddings taken from the middle and from the top of the large BERT.
\begin{table*}[th]
\caption{Summary level correlations $\rho$ (Spearman) and $\tau$ (Kendall Tau-c) of quality estimators with human experts scores. The top rows evaluation measures are reference-free, separated from the lower rows evaluation measures, which need human references. In each column the highest correlation is bold-typed. The only p-values above 0.01 in this table are p=0.03 for BERTScore-P and p=0.01 for ROUGE-2.}
\centering
\begin{tabular}{@{}lllllllll@{}}
\toprule
\textbf{measure} & \multicolumn{2}{c}{consistency} & \multicolumn{2}{c}{relevance} &
\multicolumn{2}{c}{coherence} & \multicolumn{2}{c}{fluency}\\
\midrule
{} & \bm{$\rho$} & \bm{$\tau$}
 & \bm{$\rho$} & \bm{$\tau$}
  & \bm{$\rho$} & \bm{$\tau$}
   & \bm{$\rho$} & \bm{$\tau$}\\
BLANC-AXXL & 0.200 & 0.098 & 0.246 & 0.179 & 0.127 & 0.093 & 0.115 & 0.066\\
BLANC-BLU & 0.207 & 0.102 & 0.217 & 0.156 & 0.116 & 0.085 & 0.112 & 0.065\\
(-)ESTIME-12 & \textbf{0.374} & \textbf{0.184} & 0.140 & 0.100 & 0.238 & 0.173 & 0.343 & 0.198\\
(-)ESTIME-24 & 0.358 & 0.176 & 0.117 & 0.084 & 0.187 & 0.134 & \textbf{0.363} & \textbf{0.209}\\
(-)J-Shannon & 0.193 & 0.095 & 0.406 & 0.298 & 0.289 & 0.213 & 0.125 & 0.072\\
SummaQA-F & 0.174 & 0.085 & 0.16 & 0.113 & 0.089 & 0.065 & 0.12 & 0.069\\
SummaQA-P & 0.197 & 0.097 & 0.179 & 0.127 & 0.112 & 0.082 & 0.133 & 0.076\\
SUPERT & 0.297 & 0.147 & 0.306 & 0.222 & 0.236 & 0.175 & 0.175 & 0.101\\
\hline
BERTScore-F & 0.109 & 0.053 & 0.371 & 0.273 & \textbf{0.377} & \textbf{0.277} & 0.142 & 0.082\\
BERTScore-P & 0.055 & 0.027 & 0.268 & 0.196 & 0.323 & 0.238 & 0.126 & 0.072\\
BERTScore-R & 0.164 & 0.081 & \textbf{0.423} & \textbf{0.309} & 0.345 & 0.253 & 0.12 & 0.069\\
BLEU & 0.095 & 0.047 & 0.213 & 0.153 & 0.176 & 0.128 & 0.140 & 0.080\\
ROUGE-L & 0.115 & 0.057 & 0.241 & 0.174 & 0.170 & 0.124 & 0.079 & 0.045\\
ROUGE-1 & 0.137 & 0.067 & 0.302 & 0.220 & 0.184 & 0.134 & 0.080 & 0.046\\
ROUGE-2 & 0.129 & 0.063 & 0.245 & 0.177 & 0.146 & 0.105 & 0.063 & 0.036\\
ROUGE-3 & 0.149 & 0.073 & 0.251 & 0.180 & 0.160 & 0.116 & 0.066 & 0.038\\
\bottomrule
\end{tabular}
\label{tab:summary-level}
\end{table*}
\section{Performance on human-annotated machine-generated summaries}
We used SummEval dataset \footnote{https://github.com/Yale-LILY/SummEval} \cite{Alexander2020SummEval} for comparing ESTIME with a few well known or promising evaluation measures. The part of SummEval dataset that we use consists of 100 texts, each text is accompanied by 16 summaries generated by 16 different models, making altogether 1600 text-summary pairs. Each text-summary pair is annotated (on scale 1 to 5) by 3 experts for 4 qualities: consistency, relevance, coherence and fluency. We took average of the expert scores for each quality of a text-summary pair. Each text is also accompanied by 11 human-written reference summaries, for the measures that need them.

We calculated scores of ESTIME and other measures for all the 1600 summaries, and presented their correlations with the average expert scores in Table \ref{tab:summary-level}. The measures in Table \ref{tab:summary-level} are split into the group of reference-free measures (top) and the measures requiring human-written references (bottom). 
BLANC-help \cite{Oleg2020BLANC} is calculated in two versions\footnote{https://github.com/PrimerAI/blanc\#blanc-on-summeval-dataset}, which differ by the underlying models: BLU - bert-large-uncased, and AXXL - albert-xxlarge-v2. ESTIME and Jensen-Shannon \cite{Annie2009Automatically} values are negated. 
SummaQA \cite{Thomas2015Answers} is represented by SummaQA-P (prob) and SummaQA-F (F1 score)\footnote{https://github.com/recitalAI/summa-qa}. SUPERT \cite{Yang2020SUPERT} is calculated as single-doc with 20 reference sentences 'top20'\footnote{https://github.com/yg211/acl20-ref-free-eval}. BLEU \cite{Papineni2002BLEU} is calculated with NLTK. BERTScore \cite{Tianyi2020BERTScore} is represented as F1 (BERTScore-F), precision (BERTScore-P) and recall (BERTScore-R)\footnote{https://github.com/Tiiiger/bert\_score}. For ROUGE \cite{Lin2004ROUGE} the ROUGE-L is calculated as rougeLsum\footnote{https://github.com/google-research/google-research/tree/master/rouge}.

By design ESTIME should perform well for consistency, and indeed it beats other measures in the table. Being a one-sided summary-to-text estimator of inconsistencies, ESTIME should not and does not perform well for relevance. Interestingly, ESTIME performs better than other measures for fluency, and reasonably well for coherence. 

\begin{table}[th]
\caption{System level correlation $\rho$ (Spearman) and $\tau$ (Kendall Tau-c) of quality estimators with human experts scores of consistency. Top rows show reference-free evaluation measures.}
\centering
\begin{tabular}{@{}lcccc@{}}
\toprule
\textbf{measure} & \bm{$\rho$} & \bm{$\tau$}\\
\hline
BLANC-AXXL & 0.812 & 0.617 \\
BLANC-BLU & 0.724 & 0.567 \\
(-)ESTIME-12 & 0.756 & 0.583 \\
(-)ESTIME-24 & 0.815 & 0.633 \\
(-)J-Shannon & 0.753 & 0.533 \\
SummaQA-F & 0.862 & 0.667 \\
SummaQA-P & \textbf{0.912} & \textbf{0.750} \\
SUPERT & 0.832 & 0.633 \\
\hline
BERTscore-F & -0.029 & -0.017 \\
BERTscore-P & -0.329 & -0.217 \\
BERTscore-R & 0.885 & 0.733 \\
BLEU & -0.150 & -0.017 \\
ROUGE-L & 0.376 & 0.283 \\
ROUGE-1 & 0.694 & 0.517 \\
ROUGE-2 & 0.779 & 0.600 \\
ROUGE-3 & 0.888 & 0.717 \\
\bottomrule
\end{tabular}
\label{tab:system-level-consistency}
\end{table}

In Table \ref{tab:system-level-consistency} we show correlations on system level, meaning that the scores (of automated measures and of human experts) are averaged over the 100 texts, so that each array of scores has length only 16 rather than 1600 \cite{Alexander2020SummEval}. The purpose of this would be a comparison of the summarization models. We present results for consistency in the table; for other qualities some measures have p-value higher than 0.05.  

\section{Performance on human summaries with generated subtle errors}
Machine-generated summaries, even by abstractive summarization models, generally follow the source text by frequently reproducing large spans from it. It is no surprise that the most similar context should point to the same token, thus helping ESTIME to be a good factual consistency measure. 

Human summaries are more varied in describing the source text, and it is interesting how useful can be ESTIME for evaluating them. Fundamentally, we are asking how flexible are the embeddings in understanding the context. 
In order to answer this question, we made random selection of 2000 text-summary pairs from CNN/Daily Mail dataset \cite{Karl2015Teaching}. For each human-written summary we then added the same summary modified by generated factual errors. We thus made 4000 text-summary pairs. We assigned the 'golden' scores as 1 to each clean summary, and 0 to each summary with errors. 

Our 'subtle errors' generation method is simple, heuristic-free and easily reproducible. In order to generate an error, we randomly select a whole-word token in the summary, mask and predict it by an LM model (we used bert-base-cased). We then select the top predicted candidate that is not equal to the real token, and substitute it for the real token. The resulting 'subtle errors' are indeed subtle and similar to real machine-generated mishaps and hallucinations, with the fluency preserved.

Thus, we made the evaluation task double difficult: the summaries are human-written, and the errors are subtle. For the evaluation presented in Table \ref{tab:correlations-on-generated-errors} we created 3 subtle errors in each 'score=0' summary. ESTIME is more sensitive to such subtly generated errors than other measures, as shown in Table \ref{tab:correlations-on-generated-errors}. 
All p-values in the table are less than $10^{-3}$, except $0.023$ for BLANC-AXXL, $0.002$ for Jensen-Shannon and $0.001$ for SummaQA-P.

\begin{table}[th]
\caption{Correlation $\rho$ (Spearman) and $\tau$ (Kendall Tau-c) of quality estimators with the presence of generated subtle errors in human summary. The dataset of 4000 text-summary pairs was created by random pick of 2000 test-summary pairs from CNN / Daily Mail dataset, duplicating these 2000 pairs, and by generating subtle errors in the 2000 duplicated summaries.}
\centering
\begin{tabular}{@{}lcccc@{}}
\toprule
\textbf{measure} & \bm{$\rho$} & \bm{$\tau$}\\
\hline
BLANC-AXXL & 0.036 & 0.042 \\
BLANC-BLU & 0.076 & 0.087 \\
(-)ESTIME-12 & 0.138 & 0.159 \\
(-)ESTIME-24 & \textbf{0.169} & \textbf{0.195} \\
(-)J-Shannon & 0.048 & 0.055 \\
SummaQA-F & 0.055 & 0.064 \\
SummaQA-P & 0.054 & 0.062 \\
SUPERT & 0.107 & 0.123 \\
\bottomrule
\end{tabular}
\label{tab:correlations-on-generated-errors}
\end{table}

\section{Conclusion}

We introduced ESTIME: estimator of summary-to-text inconsistency by mismatched embeddings, - a new reference-free summary quality measure, with emphasis on measuring factual inconsistency of the summary with respect to the text. In view of its good performance, we intend to use ESTIME for improving faithfulness of summarization. 

We also introduced and used for evaluation a method for generating 'subtle errors'; the method has a potential for creating consistent and realistic benchmark datasets for factual consistency.  

\bibliography{anthology,ESTIME}
\bibliographystyle{acl_natbib}

\end{document}